\documentclass{article}

 \usepackage[dblblindworkshop, final]{neurips_2025}

\usepackage[utf8]{inputenc} 
\usepackage[T1]{fontenc}    
\usepackage{hyperref}       
\usepackage{url}            
\usepackage{booktabs}       
\usepackage{amsfonts}       
\usepackage{nicefrac}       
\usepackage{microtype}      
\usepackage{xcolor}         

\usepackage{amsmath,amssymb,amsfonts}
\usepackage{textcomp}
\usepackage{graphicx}
\usepackage{amsthm}
\usepackage{subcaption}
\usepackage{mathtools}
\usepackage{multicol}

\usepackage{natbib}
\newcommand{\zspace}{\mathcal{Z}}
\newcommand{\uspace}{\mathcal{U}}
\newcommand{\vvspace}{\mathcal{V}}
\newcommand{\wspace}{\mathcal{W}}

\newcommand{\xispace}{\Xi}
\newcommand{\model}{\mathcal{M}}

\newcommand{\PP}{\mathbb{P}}
\newcommand{\QQ}{\mathbb{Q}}
\newcommand{\E}{\mathbb{E}}

\title{GenUQ: Predictive Uncertainty Estimates via Generative Hyper-Networks}
\workshoptitle{Structured Probabilistic Inference \& Generative Modeling}

\author{%
  Tian Yu Yen \\
  Sandia National Laboratories\\
  Albuquerque, NM 87123 \\
  \texttt{tyen@sandia.gov} \\
  \AND
  Reese E.~Jones \\
  Sandia National Laboratories\\
  Livermore, CA 94551 \\
  \texttt{rjones@sandia.gov} \\  
  \AND
  Ravi G.~Patel\thanks{Corresponding author} \\
  Sandia National Laboratories\\
  Albuquerque, NM 87123 \\
  \texttt{rgpatel@sandia.gov} \\
}

\begin{document}

\maketitle

\begin{abstract}
Operator learning is a recently developed generalization of regression to mappings between functions. It promises to drastically reduce expensive numerical integration of PDEs to fast evaluations of mappings between functional states of a system, i.e., surrogate and reduced-order modeling. Operator learning has already found applications in several areas such as modeling sea ice, combustion, and atmospheric physics. Recent approaches towards integrating uncertainty quantification into the operator models have relied on likelihood based methods to infer parameter distributions from noisy data. However, stochastic operators may yield actions from which a likelihood is difficult or impossible to construct. In this paper, we introduce, GenUQ, a measure-theoretic approach to UQ that avoids constructing a likelihood by introducing a generative hyper-network model that produces parameter distributions consistent with observed data. We demonstrate that GenUQ outperforms other UQ methods in three example problems, recovering a manufactured operator, learning the solution operator to a stochastic elliptic PDE, and modeling the failure location of porous steel under tension.
\end{abstract}

\section{Introduction}
Over the last few years there has been tremendous growth at the intersection of scientific machine learning (SciML), traditional scientific computing methods, and computer science algorithms.
In particular, there has been growing interest in utilizing neural networks to learn physical operators of dynamical systems, such as high-dimensional partial differential equations (PDEs) \cite{patel2018nonlinear,patel2021morp,LMC22-OL,KLL23-OL}.
These methods are referred to under the general category of operator learning.
However, in order for these operator learning approaches to achieve significant impact in scientific and engineering applications, they must have reliable and robust methods for assessing the uncertainties associated with training and deploying models.

Uncertainties from machine learning models are often placed into two categories: aleatoric uncertainties 
 arising from inherent stochasticities, e.g., in stochastic ODEs and PDEs; and epistemic uncertainties arising from lack of data.
 Ref. \cite{PMZ23-UQSciML} provides a thorough review of the types of uncertainties that can influence solutions to various dynamical systems obtained via SciML methods, such as operator learning.
In this paper, we focus on the case of aleatoric uncertainties.
In particular, we are interested in learning an operator that accurately estimates predictive uncertainty due to such variability.
That is, given a fixed input $x$, what is the probability distribution $p$ that best describes the action of a stochastic operator $\model$ acting on $x$?

Many current approaches, including Bayesian and Frequentist frameworks, rely on a likelihood function to model aleatoric uncertainty.
The stochastic operator $\model$ is parameterized by some model architecture. Optionally, a prior probability distribution can also be placed on the parameters of the model. By assuming a form for the likelihood function on the data-model relationship, the parameters of the model can be found via maximum likelihood estimation (MLE) or by maximum a posteriori (MAP) estimation \cite{PMZ23-UQSciML}. Often, the likelihood is also parameterized and optimized to more flexibly model the stochastic behavior of the system \cite{kendalluq}.
These sample parameters can then be utilized to evaluate the model and, when convolved with the likelihood function, be used to estimate the predictive uncertainty of the output of the operator.
However, these approaches can be computationally prohibitive to execute, and approximation methods are often used, such as assuming no correlation between model outputs \cite{kendalluq}.
In addition, the computational complexity of likelihood based approaches, they often also require strong assumptions about the likelihood function which may be inappropriate for assessing the type of uncertainties relevant to systems of stochastic ODEs and PDEs.

Recently, generative modeling has demonstrated success across a variety of machine learning tasks \cite{bengesi2024gm}. These models can also be interpreted as providing uncertainty quantification \cite{mimikos-stamatopoulos2024scorebased}. Score-based approaches towards generative modeling are particularly attractive because they neither require a likelihood nor involve adversarial optimization \cite{pacchiardi2022score}. Authors have  used generative models as hyper-networks to generate the weights for a neural network model, allowing one to estimate uncertainties for model predictions \cite{krueger2017bayesian,ratzlaff2019hypergan,safta2025uncertainty}. These works, however, did not provide isolated analysis of quantified aleatoric uncertainties apart from epistemic uncertainties. Moreover, instead of using score based approaches to generative modeling, they relied either on generative adversarial approaches or methods that require likelihood functions. Finally, none of these works have applied these hyper-network UQ approaches towards operator learning.

In this paper, we propose a new method of determining predictive uncertainty estimates for learned operators. We will refer to this method as GenUQ.
Our approach is to determine a set of operators whose action is consistent with the distribution of observed data by training a generative hyper-network to sample a corresponding distribution of model parameters.
The generative hyper-network can be trained at the same time as a traditional operator learning model using standard backpropagation.
In addition, the approach requires no assumptions about the likelihood of the data nor a prior distribution on model parameters (though an appropriate prior initialization may benefit training).
Surprisingly, though the parameterization of models may be high-dimensional, we show the generative hyper-network need only produce a small subset of these to produce an appropriate distribution of predictions. Therefore, GenUQ is only marginally more expensive to train and perform inference on than a deterministic model.
Though our interest is on applications in operator learning, the approach is general enough to be applied to other types of machine learning problems.



\section{GenUQ: Predictive Uncertainty from Generative Models}
Suppose we have a dataset consisting of $N$ pairs of data, $(u_i,v_i)^{i \in 1,2,\hdots,N} \in \uspace \times \vvspace$, and wish to model it by providing a function (e.g., a neural network), ${f}:\uspace\times\Theta\rightarrow\vvspace$ with appropriate parameters, $\theta \in \Theta$. In our scenario, the phenomenon that produced the dataset is stochastic in the sense that repeated experiments on the same input, $u$, from the probability measure, $\PP_\uspace(U)$, would yield  outputs from a measure, $v \sim \PP_\vvspace(V|U=u)$. For our applications, $\uspace$ and $\vvspace$ will be subspaces of $L^2$ with norms, $||\cdot||_{\uspace}$ and $||\cdot||_{\vvspace}$. We are particularly interested in operator learning based PDE surrogate models, so these spaces will be of functions over a domain, $\Omega \subset \mathbb{R}^d$. Additionally, we will not assume that we have access to multiple realizations of the phenomenon for the same input, i.e., the dataset consists of unique $u_i$'s.

Similar to Bayesian frameworks such as \cite{N12-BNNThesis,MBK21-MFBNN}, we assume that the parameters $\theta$ are random variables and can model the dataset by providing a probability measure for $ \PP_\Theta$ or a means of sampling $\theta \sim \PP_\Theta$ such that its pushforward matches the data. However, instead of following the traditional frameworks of finding a distribution for $\theta$ given the data through Bayes' Rule, we instead seek to estimate a measure $\PP_\Theta$ whose push-forward approximates the output target measure $\PP_{\wspace}(W)$. That is,
\begin{equation}
\begin{aligned}
\PP_\Theta \left( f^{-1}_u(V)\right) \equiv \widehat{\PP}_{\vvspace}(V|U=u)\approx \PP_{\vvspace}(V|U=u)
\end{aligned}
\end{equation}
where $f_u^{-1}(V)=\{\theta\in\Theta : f_\theta(u) = V\}$. This approximation should hold over any $u\sim \PP_\uspace$, so we seek a model that approximates the joint distribution, $\widehat{\PP}_{\wspace}(W) \approx \PP_{\wspace}(W)$, where  $\wspace \equiv \uspace \times \vvspace$ is also a $L^2$ space whose elements consist of $h_u(\theta) = (u,f_{\theta}(u))$. Therefore, we require consistency between the pushforward of our parameter measure and the data distribution,
\begin{align}
\PP_\Theta \left( h^{-1}(W)\right) \equiv \widehat{\PP}_{\wspace}(W)\approx \PP_{\wspace}(W)
\end{align}
where $h^{-1}(W)=\{\theta\in\Theta : h_U(\theta) = (U,f_\theta(U)) = W\}$. This approach bears similarity to other measure-theoretic frameworks for uncertainty quantification \cite{BJW18-DCIClassic,BSW25-DCIRecent}.




Note that in many applications, the uncertainty in parameters $\theta$ of a black-box model is not of primary interest: what is of interest is the accuracy and appropriateness of the approximate predictive measure $\widehat{\PP}_\wspace$. 
Therefore, we propose to use an hyper-network generative model $g_\phi$ to learn to sample  from the measure $\PP_\theta$.
Let $z\sim \PP_\zspace$ be from a measure that is straightforward to sample, such as the standard normal distribution.
We train a simple, generative feed-forward network $g_\phi:\zspace\rightarrow\Theta$ such that
\begin{align*}
    g_\phi(z) = \theta\ &\rightarrow\ f_\theta(u) = \hat{v}
\end{align*}
and we minimize the differences between the resulting distribution of $\hat{w}=(u,\hat{v})$ and $w=(u,v)$ using a discrepancy function $D$:
\begin{align}
    \min_\phi D(\widehat{\PP}_{\wspace},\PP_{\wspace}).
\end{align}
As discussed in the following section, there are many  options for learning such a sampling distribution, including normalizing flows \cite{rezende2015variational}.
In our experiments, we find that utilizing a non-invertible feed-forward neural network for $g_\phi$ is sufficient provided an appropriate discrepancy function. To maintain the computational efficiency of GenUQ, we seek a generative model for only a small, random subset of the underlying model's parameters, leaving the remainder as deterministic variables. The proportion of generated parameters, $R$, is a hyperparameter that we examine in the results section.

There are many metrics and discrepancy scores $D$ that could be utilized to estimate the differences between probability measures $\widehat{\PP}_\wspace$ and $\PP_\wspace$, such as the KL-divergence and the negative log likelihood.
We choose to utilize the energy score which is a strictly proper score in the sense that it obtains a minimum if and only if $\widehat{\PP}_\wspace = \PP_\wspace$,
\begin{equation}\label{eq:dscore}
    \begin{aligned}
    &D_{e}(\widehat{\PP},\PP):= -\frac{1}{2}\,\E_{w\sim\hat{\PP},w'\sim\PP} \left[\rho(w,w')\right] + \E_{w,w'\sim\hat{\PP}} \left[\rho(w,w')\right] \\
    &\rho(w,w')=\|w-w'\|^\beta_{\wspace} = \left(\|u-u'\|^2_{\uspace} + \|v-v'\|^2_{\vvspace}\right)^{\beta/2}
\end{aligned}
\end{equation}
where $\beta\in(0,2)$. Note that the two terms in the norm, $\|\cdot\|_\wspace$ can be weighted if $\|u\|_{\uspace}/\|v\|_{\vvspace} \neq \mathcal{O}(1)$. This score is associated with the maximum-mean discrepancy (MMD) metric for a specific choice of kernel distance \cite{BDS18-MMDGans}.
 There are a number of advantages to utilizing such a discrepancy score to train models.
See \cite{gneiting2007strictly,pacchiardi2022score} for details.
In particular, one of the advantages of using this type of discrepancy score is that explicit knowledge of the underlying distributions is not required to approximate the energy score (as with other MMD metrics), i.e., the energy score can be approximated via Monte Carlo from finite samples of the two comparison distributions $\PP$ and $\QQ$. 
Indeed, we have samples from the joint distribution $\{w_i=(u_i,v_i)\}\sim\PP_{\wspace}$ in practice.
Thus, we can train the hyper-network generative model with the following criteria:
\begin{equation*}
\begin{matrix*}[l]
    z_{ij} \sim \PP_\zspace: &\text{samples from generating distribution}\\
    \theta_{ij} =g_\phi(z_{ij}): &\text{model parameter samples}\\
    \hat{v}_{ij} = f_{\theta_{ij}}(u_i): &\text{model outputs}\\
    \hat{w}_{ij} = (u_i,\hat{v}_{ij}): &\text{predicted samples}\\
    \min_\phi  \widetilde{D}_e  (\widehat{\PP}_{\wspace},\PP_{\wspace}): & \text{objective function}
\end{matrix*}
\end{equation*}
where
\begin{align*}
\widetilde{D}_e (\widehat{\PP}_{\wspace},\PP_{\wspace}) = -\frac{1}{2mn}\,\sum_i\sum_j \rho\left(\hat{w}_{ij},w_{i}\right) \\
\quad + \frac{1}{m(n-1)}\,\sum_{i}\sum_{j\neq j'} \rho\left(\hat{w}_{ij},\hat{w}_{ij'}\right).    
\end{align*}
where $m$ the number of samples in the training set and $n_z$ is the number of model predictions and a tunable optimization hyperparameter. Although we focus on operator learning examples in the results section, we note that GenUQ is not limited to operator learning and is applicable to other regression tasks, e.g., $\uspace=\mathbb{R}^n$ and $\vvspace=\mathbb{R}^m$.

\section{Related Works}
Bayesian approaches for uncertainty quantification are very popular, especially in the scientific machine learning community.
Among them, variational inference is commonly chosen because it enables inference for posterior parameters in situations where performing MCMC sampling is challenging \cite{G11-VI4NN,BKM17-VIReview}.
Nonetheless, classical variational inference approaches can be limited by assumptions on the class of approximation densities used to approximate the posterior (e.g., mean-field approximations).
To overcome this limitation, strategies such as normalizing flows have been proposed to generate a more flexible class of approximation densities for use in the variational inference approach \cite{RM15-VIwNF}.
Our approach is in a similar spirit to this type of approach but differs in the mathematical framework underlying the approach.
While variational inference also seeks a distribution over model parameters, the posterior distribution it attempts to approximate is assumed to be a convolution of the data likelihood and prior distribution.
Instead, we seek a distribution over the model parameters whose push-forward measure is consistent with the distribution of the data, i.e., via the discrepancy score \eqref{eq:dscore}.
These differences in objectives can lead to significantly different 
mathematical consequences--see \cite{BJW18-DCIClassic} for details.
Our approach is most similar to \cite{LGM24-fuse}, which learns a solution to an inverse problem (in addition to the forward operator) to characterize the stochastic inputs to a PDE.
They associate every input $u\in\uspace$ with a specific stochastic parameterization $\xi\mid u\in \xispace$, and then learn the appropriate mapping $f_\theta:\xispace\rightarrow\vvspace$.
On the other hand, we assume that $\xi$ is independent of $u$ and directly influences the output $v$ through the operator.
Thus, our method, GenUQ, focuses on a forward problem directly relating $\uspace$ and $\vvspace$ (via $f_\theta:\uspace\rightarrow\vvspace$), bypassing the need to solve an extra inverse problem.


\section{Results} \label{sec:results}

We evaluate and compare GenUQ to other UQ methods in three examples. Our first two examples utilize neural operators as the underlying model, $f$, while the final example uses a convolutional neural network. For each example, we have a dataset of input/output function pairs $(u,v)$, that we partition into training/validation/test sets with a 60\%/20\%/20\% split. We train all models with the Adam optimizer using a stepped learning rate schedule: $(10^{-3},10^{-4},10^{-5},10^{-6})$, with 400 epochs per learning rate. We track the validation loss over the training process and report the model with the lowest validation loss. All experiments were performed on a 80GB A100 NVIDIA GPU.

Throughout these examples, we compare to 5 other UQ methods.  The first three are maximum likelihood estimation based methods where we assume Gaussian error with heteroscadastic, but uncorrelated noise in a mean field variational inference model (VI), heteroscadastic noise with full covariance structure (CoV), and normalizing flow conditioned on the input function (NF). While VI provides both epistemic and aleatoric uncertainties, we include it due to its ubiquity in machine learning. For CoV, we predict a full covariance matrix from the neural operator. For NF, we use masked autoregressive flow \cite{papamakarios2017masked} from the error distribution to a normal distribution using the FlowJax library with default options \cite{ward2023flowjax}.

 The forth and fifth methods we compare to GenUQ are a dropout (DO) architecture  and a generative model that lacks the hyper-network and maps the input function concatenated with noise to the stochastic output (Generative). Both architectures are trained using the same energy score in \eqref{eq:dscore}

\subsection{ELU Operator learning example}\label{sec:elu}

\begin{figure}
    \centering
    \includegraphics[width=.8\linewidth]{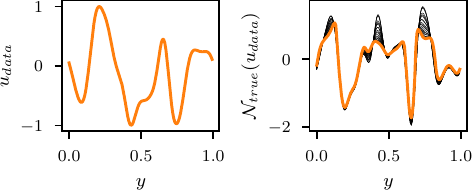}
    \caption{\textit{(Left)} Sample input  function in training data. \textit{(Right, orange)} Sample output function in training data. \textit{(Right, black)} Actions of resampled operators on same input function. }
    \label{fig:resampled_elu}
\end{figure}

We compare GenUQ to three other UQ methods in recovering a manufactured stochastic operator from its action on random smooth functions. The operator we use is the composition between the exponential linear unit (ELU) point-wise non-linearity \cite{clevert2016elu} and a differentiation,
\begin{equation}\label{eq:elu}
\begin{aligned}
    &\mathcal{N}(u) = \partial_x f \\
    &f(x) = \mathrm{ELU}(u(x)+\alpha) -\alpha\\
    &\alpha \sim U[0,1]
\end{aligned}
\end{equation}
To generate a set of input functions, $u_i$, we generate smooth random functions on a periodic 1D domain, $[0,2\pi]$, by sampling a Gaussian process with a mean of zero and a covariance kernel, $K(x,x')=\exp(4\cos(2\pi(x-x')))$. For each input function, $u_i$, we sample an operator, $\mathcal{N}_i$, as per \eqref{eq:elu}, and apply it to $u_i$ to generate $v_i = \mathcal{N}_i(v_i)$. We try to recover the stochastic operator in \eqref{eq:elu} using function pairs $\{u_i,v_i\}^{i=1,\hdots,N}$. Our full dataset consists of $N=2048$ sample pairs.
In Figure~\ref{fig:resampled_elu} we show a pair of input and output functions. Additionally, we show 20 samples of the action of the stochastic operator on the same input function to qualitatively show the stochasticity of the operator.

Our underlying model is the operator learning method, Modal Operator Regression for Physics (MOR-Physics)\cite{patel2018nonlinear,patel2021morp},
    ${f}_\theta (u) = \mathcal{F}^{-1}g_{\theta_g}(\kappa) \mathcal{F} h_{\theta_h}(u)$,
where $\mathcal{F}$ is the Fourier transform, $\kappa$ is the wavenumber, and $g$ and $h$ are neural networks. For both neural networks, we use a depth of 6 and width of 32 and the GELU activation function. For the generator network, we use a four layer network of widths, $(10,20,30,40)$, and the same activation function.

\begin{figure*}[t!]
    \centering
    \begin{subfigure}[t]{0.5\textwidth}
        \centering
        \includegraphics[width=0.8\linewidth]{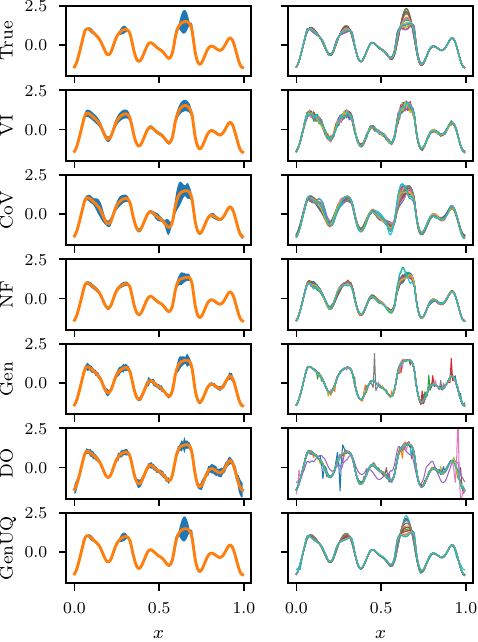}
        \caption{Full input range}
    \end{subfigure}%
    ~ 
    \begin{subfigure}[t]{0.5\textwidth}
        \centering
        \includegraphics[width=0.8\linewidth]{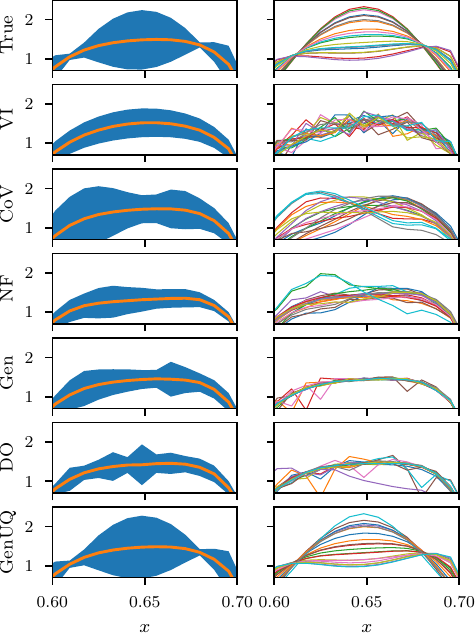}
        \caption{Focused input range}
    \end{subfigure}
    \caption{Comparison of UQ methods for recovering the 1D operator in \eqref{eq:elu}. \textit{(Left)} the predictive mean \textit{(orange)} and 95\% confidence intervals \textit{(blue)} of the action of each method on a test function. \textit{(Right)} Actions of resampled operators on same test function. Qualitatively, the learned GenUQ operator better matches the true operator. Energy distances between predictions and data, VI: 0.0265, CoV: 78.4386, NF: 0.1215,   Gen: 0.2122, DO: 0.1994,  GenUQ: 0.0020. }
    \label{fig:elu}
\end{figure*}

In addition to training a GenUQ model on this dataset, we train the five other UQ methods discussed above. 
After training the models, we generate new input functions from the same Gaussian process to compare model performance. Figure~\ref{fig:elu} shows the comparison between the true operator \eqref{fig:resampled_elu}, the VI, CoV, NF, DO, Gen, and GenUQ operators. We find that the GenUQ operator better captures the stochasticity of the true operator than the other methods and has lower energy distance. The 95\% confidence intervals provided by GenUQ more closely follow those of the true operator compared to than the others. Moreover, the actions for the GenUQ operator remain smooth while the error model in VI that assumes heteroscadastic but independent Gaussian noise loses the correlation structure found in the data. Similarly, Gen and DO both produce noisy predictions. CoV and NF is able to capture smooth functions but the Gaussian restriction for CoV and optimization challenges for both methods led to to poor quality UQ. 

\begin{figure}
    \centering
        \begin{subfigure}[t]{\textwidth}
        \centering
        \includegraphics[width=0.6\linewidth]{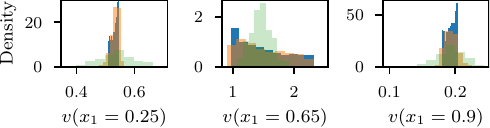}
        \caption{Histograms of actions at single point.}
        \label{fig:elu_hist}
    \end{subfigure}%
    \\
    \begin{subfigure}[t]{\textwidth}
        \centering
        \includegraphics[width=0.6\linewidth]{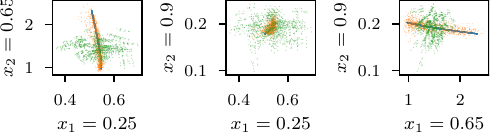}
        \caption{ Pair-wise values of actions.}
        \label{fig:elu_scatter}
    \end{subfigure}
    \caption{ Fine grained evaluation of GenUQ in ELU operator learning example. Stochastic operators' actions on test input function in Figure~\ref{fig:elu}.  \textit{(blue)} Test data. \textit{(green)} VI.  \textit{(orange)} GenUQ. }
\end{figure}

In Figure~\ref{fig:elu_hist}, we look at the distribution of the action, $v$, at three different points in the domain. For clarity, we only compare GenUQ to VI. We observe that GenUQ produces distributions that are closer to the data than VI. In Figure~\ref{fig:elu_scatter}, we look more closely at the correlation structure for the actions of the three operators by plotting sample points, $(v_t(x_0),v_t(x_1))$ for $x_0=0.6$ and varying $x_1$. We find that VI has weak correlations throughout while GenUQ matches closely the correlation structure of the true operator. Additionally, we plot the histograms of $v_t(x_1)$ for the three operators for varying $x_1$ and find GenUQ again to more closely match the true operator than VI. GenUQ, however, does not exactly match the true operator and produces distributions for the actions that are more diffuse than the sharp distributions from the true operator.


We assess the sensitivity of the GenUQ model to the random generator dimension $d$ and the number of model parameters treated as stochastic $S$. We measured the quality of the predicted distributions for each $d$ and $S$ using the energy distance, which is equivalent to the energy score \eqref{eq:dscore} plus $\E_{w,w'\sim\PP} \left[\rho(w,w')\right]$. We also checked the accuracy of the predicted sample means in the $L^2$-norm. Table~\ref{tab:psweep} shows the results. For this example, only a small proportion of the total number of parameters needs to be utilized to reproduce reasonable actions of the learned stochastic operator. This is very likely due to the fact that the stochasticity of the true operator is governed by a single random variable, $\alpha$. On the other hand, treating too many model parameters as stochastic leads to a lack of convergence. We hypothesize that the model needs to be reasonably calibrated to the data in order for the model to converge to an appropriate solution. When too many parameters are stochastic, the non-uniqueness of model parameters leads to degeneracy of the learning process.

\begin{table*}[t]
    \centering
    \begin{tabular}{c|cccc}
         $S=(\%$ of $M)$  &  \multicolumn{4}{c}{$d=$ ($\%$ of $S$)}   \\
         & 25 & 50 & 75 & 100 \\
        \hline
        0.1  & 0.005 (2.2e-4) & 0.004 (2.0e-4) & \bf{0.002 (1.6e-4)} & 0.005 (2.1e-4)\\
        0.4  & 0.003 (1.9e-4) & 0.004 (1.9e-4) & 0.003 (1.8e-4) & 0.002 (1.7e-4)\\
        1.6  & 0.005 (5.2e-4) & 0.005 (1.9e-4) & 0.003 (1.7e-4) & 0.008 (2.8e-4)\\
        6.4  & 0.122 & DnC & DnC & DnC\\
        25.6 & DnC & DnC & DnC & DnC\\
    \end{tabular}
    \caption{Shows the GenUQ model performance for different numbers of stochastic parameters $S$ (as a proportion of $M$) and generating dimension $d$ (as a proportion of $S$); the total number of model parameters was $M=16,131$. Performance measured in the energy distance \eqref{eq:dscore} and differences between sample means in $L^2$-norm in parentheses. DnC: did not converge.}
    \label{tab:psweep}
\end{table*}

\subsection{Stochastic Poisson equation solution operator}\label{sec:poisson}

\begin{figure}
    \centering
    \includegraphics[width=.6\linewidth]{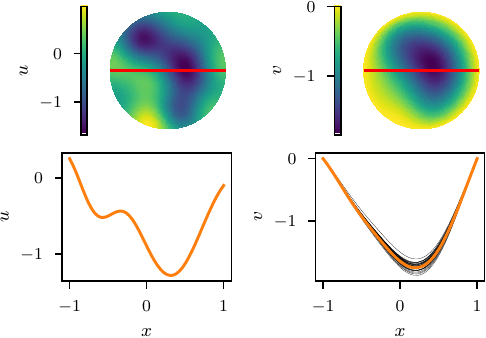}
    \caption{\textit{(Top Left)} Sample input function from training set. \textit{(Top Right)} Sample action on input function from training set. \textit{(Bottom Right)} Trace of sample input function along the red line in the corresponding contour plot. \textit{(Bottom Left, orange)} Trace of sample output function along the red line. \textit{(Bottom Left, black)} Action of multiple realizations of solution operator on fixed input function. }
    \label{fig:poisson_data}
\end{figure}

In our next example, we compare GenUQ to VI for recovering the solution operator for a stochastic nonlinear Poisson equation. We consider the following PDE on the unit disk in 2D,
\begin{equation}\label{eq:poisson}
    \begin{matrix}
        \nabla \cdot a (\nabla v) = u & ||x||_2 < 1 \\
        u = 0 & ||x||_2 = 1 \\
        a \sim \mathcal{A} \quad u \sim \mathcal{U}
    \end{matrix}
\end{equation}
We select for $\mathcal{U}$ a Gaussian process, $\mathcal{G}[0,K(x,x')=0.04\exp(6.25||x-x'||_2^2)]$. To maintain the well-posedness of the PDE, $a: \mathbb{R}^2\rightarrow \mathbb{R}^2$ must be a monotone function \cite[Chapter~9]{evans2022partial}, so $\mathcal{A}$ must be a distribution over monotone functions. We do not explicitly construct $\mathcal{A}$, but instead sample $\mathcal{A}$ by randomly initializing a one hidden layer of width 2 monotone neural network with hard sigmoid activations. To generate the training, validation, and test data, we numerically solve \eqref{eq:poisson} using a finite element solver with P1 basis functions on a triangular mesh with characteristic mesh size, $h=0.025$. 
We generate a dataset of $10000$ samples.
In Figure~\ref{fig:poisson_data} we show sample input and output function pairs from the test dataset. 

We will refer to the stochastic solution operator for \eqref{eq:poisson} provided by the finite element solver as $N_{true}$ and seek a neural operator, $N$, that emulates it. We  parameterize the neural operator with the POD-DeepONets architecture with hard constrained boundary conditions \cite{Lupod-don2022}. We construct a truncated POD basis of dimension $d=200$ from the training input functions, and project the input functions onto this basis to produce the POD coefficients, $c$. The POD-DeepONets architecture has the following form,
\begin{equation}
\begin{aligned}
    &f_\theta(u)(x) = g(x) (b_{\theta_b}(c) \cdot t_{\theta_t}(x))\\
    &g(x) = 1-x^2
\end{aligned}
\end{equation}
where $t$ and $b$ are trunk and branch neural networks of depth 5 and width 128. See \cite{Lupod-don2022} for further details on this architecture. 

\begin{figure}
    \centering
    \begin{subfigure}[t]{\linewidth}
    \centering
    \includegraphics[width=0.6\linewidth]{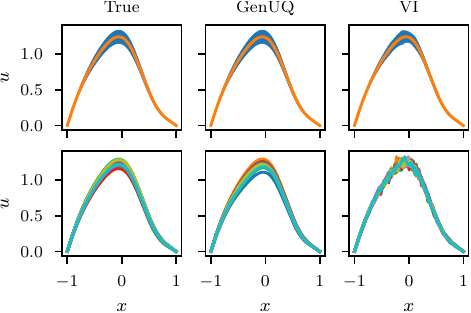}
    \caption{Full input range} 
    \end{subfigure}
    \\
    \begin{subfigure}[t]{\linewidth}
    \centering
    \includegraphics[width=.6\linewidth]{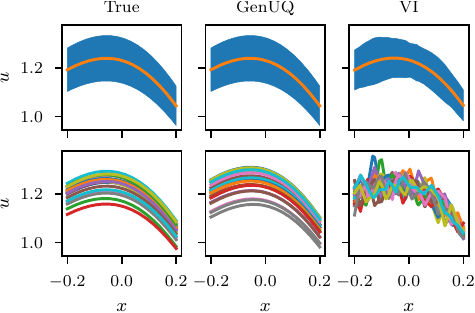}
    \caption{Focused input range} 
    \end{subfigure}
    \caption{Visual comparison of VI and GenUQ for recovering the stochastic Poisson equation. between predictions and data, VI: 0.2643 and GenUQ: 0.0780}
    \label{fig:poisson}
\end{figure}

\begin{figure}
    \centering
        \begin{subfigure}[t]{\textwidth}
        \centering
        \includegraphics[width=0.6\linewidth]{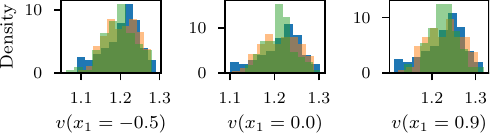}
        \caption{Histograms of actions at single point.}
        \label{fig:poisson_hist}
    \end{subfigure}%
    \\
    \begin{subfigure}[t]{\textwidth}
        \centering
        \includegraphics[width=0.6\linewidth]{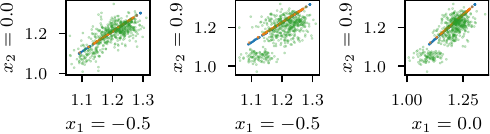}
        \caption{ Pair-wise values of actions.}
        \label{fig:poisson_scatter}
    \end{subfigure}
    \caption{ Fine grained evaluation of GenUQ in Poisson example. Stochastic operators' actions on test input function in Figure~\ref{fig:poisson}.  \textit{(blue)} Test data. \textit{(green)} VI.  \textit{(orange)} GenUQ. }
\end{figure}

We train VI  and GenUQ models with the same underlying architecture. In Figure~\ref{fig:poisson}, we evaluate both models on a test input function. As in the previous section, we find that GenUQ more closely matches the test data than VI. Again, VI produces noisy predictions due to the poorly specified likelihood function. The energy distance for the GenUQ model is also lower. In Figure~\ref{fig:poisson_hist} we plot histograms for the predictions of the three operators for an test input function at three points and find that GenUQ matches the data more closely than VI.  Figure~\ref{fig:poisson_scatter}, we find that the pair-wise values for GenUQ at these points also more closely matches the test data. 

\subsection{Porosity Example}

For our last example, we apply GenUQ to learn the mapping between the porosity of a steel sample and the location of its failure under strain. We use high fidelity data from \cite{Khalil2021porosity} which consists of $0.75\mu$m$\times0.75\mu$m$\times4\mu$m volumetric fields for the porosity of the material, $u$, and damage at failure, $v$. The damage field is in the range $[0,1]$ where it takes a value of $1$ at the point of failure. The porosity of the material was sampled from a Karhunen-Loeve process. For each porosity field, the authors conducted a finite element simulation of the material under strain until ductile failure. See \cite{Khalil2021porosity} for further details on the simulation.  The total number of simulations in the dataset is $9457$ which we partition into training, test, and validation datasets. 

A deterministic model for a mapping between porosity, $u$, and failure is challenging to obtain due to the extremely sensitive and highly localized nature of the map. Notably, the location of failure is often near a pore, but it is difficult to predict near which pore or cluster of pores the material will fail at. Additionally, the strain required break the material has high sample variance. Furthermore, each finite element simulation utilized approximately 16 CPU-hours. An accurate deterministic model would need to be inordinately complex to capture the behaviour in this dataset. Alternatively, we can recast the problem as one of predicting aleatoric uncertainty, i.e., given a sample porosity field, predict a distribution of failure locations. In this context, the true failure point would lie in a high density region in the support of a simpler, stochastic model's predictions. 

We compare a deterministic and GenUQ model in predicting the failure location. For both models, we use the same underlying convolutional neural network (ConvNet) architecture to find a mapping between the porosity, $u$, and damage, $v$, training on pairs of fields from the training set. The ConvNet consists of 7 blocks of 3x3 convolution, batch normalization, and GELU activation. The convolutions in the blocks have filter sizes, (4,8,16,16,8,4,1). To the find failure location, we find the location of the maximum damage predicted. In Figure~\ref{fig:porosity} in the appendix we show a sample pair of porosity and damage fields from the test set and corresponding damage predictions from the trained models. In Table~\ref{tab:porosity}, we compare statistics from test data predictions from the two models. Given enough samples from the generative model, it is able to successfully predict the failure location and damage field. The deterministic model, in contrast has very poor accuracy and high energy distance. Notably, even with only a single prediction, the generative model still outperforms the deterministic model. While this observation will require further investigation, we suspect that GenUQ may have a regularizing effect in addition to providing UQ.

\begin{table*}[t]
    \centering
    \begin{tabular}{c|c|c|c|c|c|c}
           &      Deterministic          &\multicolumn{5}{c}{GenUQ} \\
           &      &  $n_z$ = 1 & $n_z$ =4 & $n_z$ = 16 & $n_z$ = 64 & $n_z$ = 256 \\
           \hline
       Acc        & 64.24 &  63.28 & 77.35 & 84.56 & 89.84 & 93.41 \\
       $\ell_2$   & 48.94 &  42.47 & 30.34 & 23.15 & 16.62 & 11.86  
    \end{tabular}
    \caption{Metrics for porosity damage problem. Acc refers to accuracy and $\ell_2$ refers to relative $\ell_2$ error. Both values are reported as percentages. For GenUQ, the best prediction out of $n_z$ are reported. GenUQ outperforms the deterministic model even for $n_z=1$.}
    \label{tab:porosity}
\end{table*}


\section{Conclusion}

In this work, we develop a UQ strategy for modeling aleatoric uncertainty with a focus on neural operators. Stochastic nonlinear operators can produce data with complicated error structures that are difficult to model with traditional likelihood based approaches. We develop an alternative approach that avoids constructing a likelihood by introducing an hyper-network generative model that produces model parameters consistent with the data distribution after training on an energy score. This approach outperforms standard UQ approaches in a variety on operator learning tasks. While our method is capable of quantifying aleatoric uncertainty for a variety of problems, it has limited capacity for quantifying epistemic uncertainty. Additionally, while the method was successful in the three examples we studied, it is unclear how well it will perform for other tasks. Future work will focus on extending this method to better quantify epistemic uncertainty and further applications.

\section*{Acknowledgment}

Sandia National Laboratories is a multimission laboratory managed and operated by National Technology and Engineering Solutions of Sandia, LLC., a wholly owned subsidiary of Honeywell International, Inc., for the U.S. Department of Energy’s National Nuclear Security Administration under contract DE-NA-0003525. This paper describes
objective technical results and analysis. Any subjective views or opinions that might be expressed in the paper do not
necessarily represent the views of the U.S. Department of Energy or the United States Government.

SAND Number: SAND2025-12223C

\bibliography{biblio}


\newpage

\appendix

\section{Qualitative evaluation of porosity fields predicted by GenUQ model}

\begin{figure}[h!]
    \centering
    \includegraphics[width=.7\linewidth]{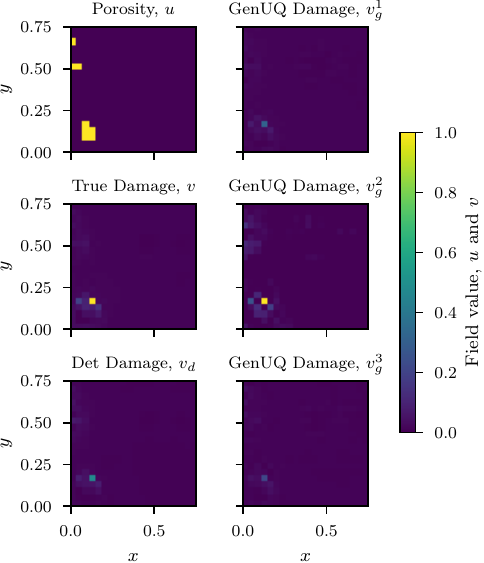}
    \caption{\textit{(Top Left)} slice of a sample porosity field from the test set at $z = 2.35$ \textit{(Middle Left)} Corresponding slice of damage field showing failure at $(x,y)=(0.1125,0.15)$. \textit{(Bottom Left)} Damage field predicted by deterministic model at same slice. Damage appears attenuated at this location. Failure point predicted incorrectly at $(x,y,z)=(0.0188, 0.3563, 0.3)$. \textit{(Left)} Three samples of damage fields predicted by GenUQ. GenUQ will occasionally predict the correct damage behavior and failure location at the cost of also sampling erroneous behaviors.}
    \label{fig:porosity}
\end{figure}

\end{document}